\documentclass{article}

    \PassOptionsToPackage{numbers, compress}{natbib}

    \usepackage[final]{neurips_2024}

\usepackage[utf8]{inputenc} 
\usepackage[T1]{fontenc}    
\usepackage{hyperref}       
\usepackage{url}            
\usepackage{booktabs}       
\usepackage{amsfonts}       
\usepackage{nicefrac}       
\usepackage{microtype}      
\usepackage{xcolor}         
\usepackage{graphicx}

\usepackage{algorithm}
\usepackage{algorithm}
\usepackage{algorithmicx}
\usepackage[noend]{algpseudocode}
\usepackage{cleveref}
\usepackage{multirow}
\usepackage{array}
\usepackage{subcaption}
\usepackage{amssymb}
\usepackage{enumitem}
\usepackage{booktabs}  
\usepackage{longtable} 

\title{Enhancing Long Context Performance in LLMs Through Inner Loop Query Mechanism}

\author{%
  Yimin Tang$^{1,2}$, Yurong Xu$^{1}$, Ning Yan$^{1}$, Masood Mortazavi$^{1}$ \\
  \texttt{\{ytang4,yurong.xu,yan.ningyan,masood.mortazavi\}@futurewei.com} \\
  $^1$Futurewei Technologies Inc, Santa Clara, 95050 \\
  $^2$University of Southern California, Los Angeles, 90089 \\
}

\begin{document}

\maketitle

\begin{abstract}
Transformers have a quadratic scaling of computational complexity with input size, which limits the input context window size of large language models (LLMs) in both training and inference. Meanwhile, retrieval-augmented generation (RAG) besed models can better handle longer contexts by using a retrieval system to filter out unnecessary information. However, most RAG methods only perform retrieval based on the initial query, which may not work well with complex questions that require deeper reasoning. We introduce a novel approach, Inner Loop Memory Augmented Tree Retrieval (ILM-TR), involving inner-loop queries, based not only on the query question itself but also on intermediate findings. At inference time, our model retrieves information from the RAG system, integrating data from lengthy documents at various levels of abstraction. Based on the information retrieved, the LLM generates texts stored in an area named Short-Term Memory (STM) which is then used to formulate the next query. This retrieval process is repeated until the text in STM converged. Our experiments demonstrate that retrieval with STM offers improvements over traditional retrieval-augmented LLMs, particularly in long context tests such as Multi-Needle In A Haystack (M-NIAH) and BABILong.
\end{abstract}

\section{Introduction}

Large Language Models (LLMs) have demonstrated a powerful ability to handle almost all kinds of NLP tasks with impressive performance~\cite{brown2020language,achiam2023gpt,team2023gemini}. As the size of LLMs increases, they tend to perform better and store more informative knowledge within their parameters~\cite{kaplan2020scaling,petroni2019language}. LLMs can also be further fine-tuned on downstream tasks~\cite{xu2021raise}. However, the length of the input window in LLMs is constrained by the quadratic computational complexity of the self-attention mechanism, which is a fundamental structure in these models~\cite{vaswani2017attention}. An alternative approach to processing longer contexts is to split large quantities of text into chunks and index these chunks as vectors in a separate information retrieval system~\cite{lewis2020retrieval,asai2023self,izacard2022few}. Since the retrieval system can filter out unnecessary information, the LLM can handle user questions with long raw data within a limited context window. Additionally, this approach provides easier interpretability and provenance tracking compared to the opaque and unexplainable parameters within LLMs~\cite{akyurek2022towards}.

However, existing retrieval-augmented approaches also have shortcomings. The issue we aim to address is that most existing methods retrieve only a few text chunks that are directly related to user queries, which limits the ability of LLMs to produce deeper answers to questions. This is particularly relevant when it comes to fully understanding or integrating knowledge from multiple parts that may not be directly related to the user's questions, such as understanding foreshadowing in novels or inferring the identity of the killer in detective fiction. When humans perform such tasks, we often have impressions in our minds related to unusual facts that may connect to our questions, indicating that memory plays a crucial role in advanced comprehension skills~\cite{blum2022theory,lecun2022path}. 
Long-term and short-term memory are essential aspects of human-like intelligence, particularly in maintaining an understanding of long contexts, such as lifelong conversations where recalling past interactions is crucial for rapport building and long context comprehension. 
Some interesting works have been explored~\cite{wilmot2021memory,zhong2024memorybank} and implemented in real products, such as memory in medical applications~\cite{zhang2024llm} and ChatGPT's memory capabilities~\cite{openai2024memory}. 
However, these works still rely on a single retrieval query in the memory system and lack a mechanism that can automatically gather more information based on the current context, making it difficult to accurately answer user questions.

To address this, we designed an inner-loop mechanism that allows retrieval to be conditioned not only on the initial query but also on the current information obtained.  Our system, Inner Loop Memory Augmented Tree Retrieval (ILM-TR), leverages the retrieved information to generate intermediate answers, which are then used to formulate subsequent queries. This inner-loop query continues until the answer converges or the query limit is reached. This mechanism can effectively answer complex questions in long context scenarios.

Our main contributions are as follows: 
1) We propose a novel summarization method which not only summarizes the chunked texts but also outputs all surprising facts within the content.
2) We propose an inner-loop mechanism which refines the retrieval process by conditioning it not only on the user's initial query but also on the evolving information gathered during the retrieval process. Our approach improves comprehension in long-context scenarios by enabling more effective information retrieval and interpretation.
3) we test the ILM-TR system on standard long-context benchmarks such as Multi-Needle In A Haystack (M-NIAH)~\cite{kamradt2024needle}, and BABILong~\cite{kuratov2024babilong} with Llama3 as our summary model and answer model. Experimental results demonstrate that our method outperforms baseline RAG method and maintains robust performance, with no significant degradation even as the context length scales up to 500k tokens.

\section{Related Work}

\subsection{Large Language Models (LLMs)}

LLMs such as the GPT~\cite{brown2020language,achiam2023gpt}, Gemini~\cite{team2023gemini,geminiteam2024gemini15unlockingmultimodal}, and Claude~\cite{anthropic2024claude3} series have made remarkable strides across a broad spectrum of tasks and have increasingly become daily assistants for many people. However, the closed-source nature of these models prohibits researchers and companies from studying the inner mechanisms of LLMs and building domain-adapted applications. Consequently, many open-source LLMs have emerged in the community, such as Llama~\cite{touvron2023LLaMa,dubey2024LLaMa}, ChatGLM~\cite{glm2024chatglm}, and Mistral~\cite{jiang2023mistral}. However, these models still have limited context windows, usually capped at 8k tokens, due to the complexity of self-attention and position encoding. The limited context windows imply that they lack long-term memory capabilities, which could be enhanced by integrating RAG systems with memory structures.

\subsection{Retrieval-Augmented Generation (RAG)}

Retrieval-Augmented Generation (RAG) uses retrieved tokens from long context raw data to extend the input window size of LLMs. The original RAG~\cite{lewis2020retrieval} integrates pre-trained sequence-to-sequence models with a neural retriever. ~\cite{min2021joint} introduced the Joint Passage Retrieval (JPR) model, which employs a tree-decoding algorithm to handle passage diversity and relevance in multi-answer retrieval. Dense Hierarchical Retrieval (DHR) and Hybrid Hierarchical Retrieval (HHR) represent advancements in retrieval accuracy by combining document-level and passage-level retrievals and integrating sparse and dense retrieval methods, respectively~\cite{liu2021dense,arivazhagan2023hybrid}. RAPTOR~\cite{sarthi2024raptor} utilizes clustering and summarizing of text chunks, constructing a tree with varying levels of summarization from the bottom up, enabling multiple levels of understanding of long contexts. Nevertheless, these approaches still depend on a single retrieval query and do not include a mechanism for automatically acquiring additional information based on the evolving context, which makes it challenging to provide precise answers to user questions.

\begin{figure*}[htb]
\centering
\includegraphics[width=\textwidth]{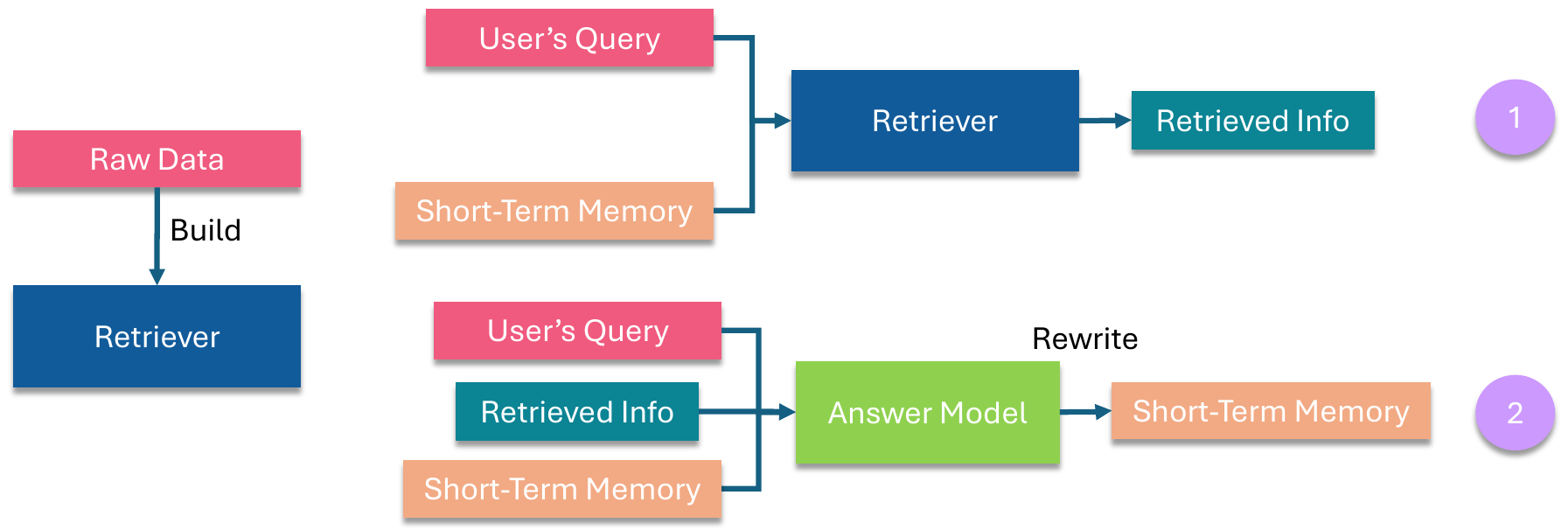} 
\caption{An Overview of \textbf{ILM-TR} Method. 
\textbf{Raw Data} consists of tokens from the user, which could include conversation history, novels, or any other content the user wants the LLMs to process. \textbf{User's Query} refers to the tokens provided by the user, such as questions or task descriptions. \textbf{Retriever} can be any retrieval method, such as sentence-based RAG or tree-structured RAG. \textbf{Retrieved Info} is the result produced by the retriever. \textbf{Short-Term Memory} is a storage area for a limited number of tokens, which is overwritten at each iteration of the inner-loop query. \textbf{Answer Model}(LLMs) will processe information from the retriever, the previous short-term memory, and the user's query. The purple circles represents the order of steps in the inner-loop query.
}
\label{fig:overview}
\end{figure*}

\subsection{Memory Mechanisms}

Efforts have been made to improve the memory capabilities of neural models. Memory-augmented networks (MANNs)~\cite{meng2018dialogue}, such as Neural Turing Machines (NTMs)~\cite{graves2014neural}, enhance the memory capacity of neural networks by incorporating an external memory matrix, allowing them to manage tasks that require the storage and manipulation of information over extended periods. ~\cite{schuurmans2023memory} demonstrates that LLMs with external memory can simulate Turing Machines. Although promising, these approaches have yet to fully address the need for a dependable and adaptable memory function in LLMs. Some studies have focused on long-range conversations~\cite{xu2021beyond,xu2022long}, but these are typically limited to a few conversational rounds, falling short of supporting long-term AI companions. Additionally, these models often struggle to create detailed user profiles and lack a human-like memory updating mechanism, both of which are essential for enabling more natural interactions.

\section{Method}

\begin{figure*}[htb]
\centering
\includegraphics[width=\textwidth]{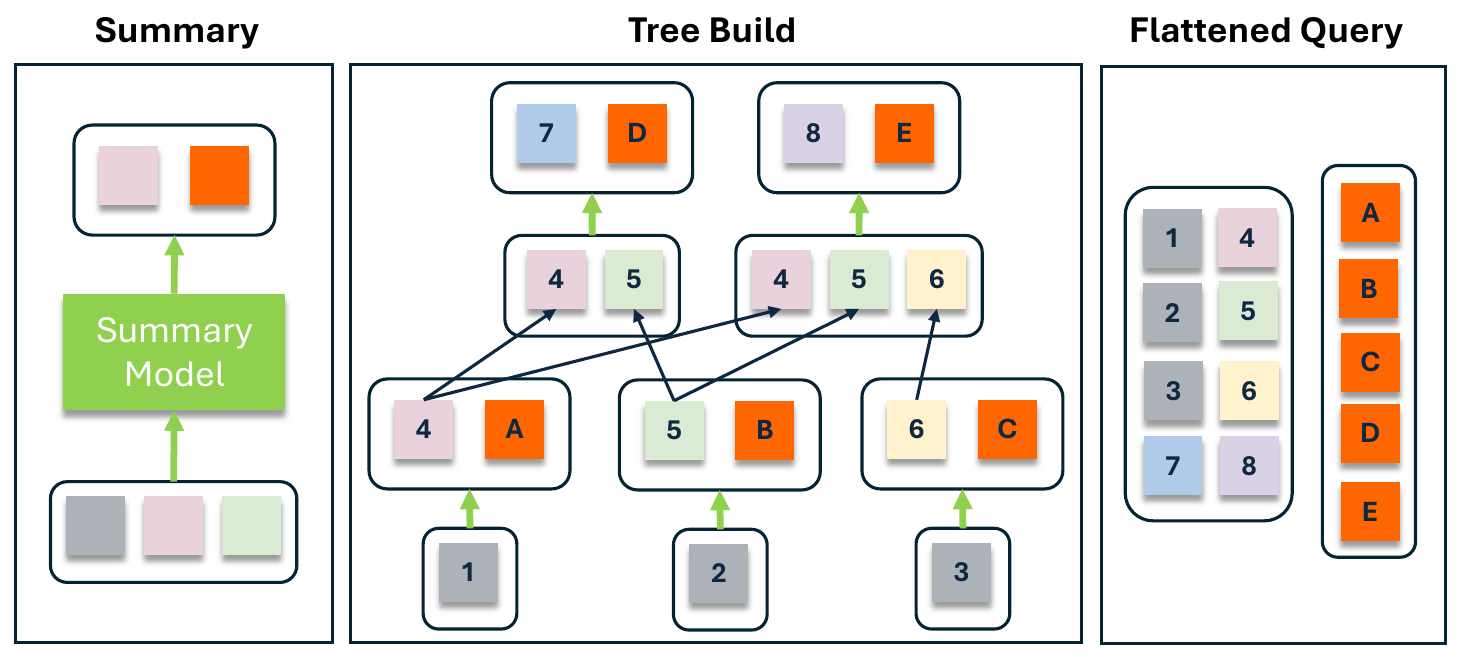} 
\caption{The \textbf{ILM-TR} Retriever: The orange square represents the surprising information, the grey color represents the original text, and other colors represent the summary information. The summary model will extract information from the provided tokens. During the tree-building process, the summary information will be grouped using a clustering algorithm, and then each group will be summarized together to generate a higher-level summary and surprising information. In the query process, all squares in the tree will be stored in a table, and the best fit will be returned based on vector distance from the query text.
}
\label{fig:overview2}
\end{figure*}

Our Inner Loop Memory-Augmented Tree Retrieval (ILM-TR) method, as shown in Fig.~\ref{fig:overview}, contains two parts: retriever and inner-loop query. For the retriever part, we primarily use the RAPTOR's tree-build method~\cite{sarthi2024raptor}. The retriever first segments the raw data into short, contiguous text chunks of a certain length. If a sentence exceeds the length limit, it will be moved to the next chunk. After splitting, an summary model is used to summarize each chunk. However, unlike typical summarization methods such as RAPTOR, our model produces two kinds of summaries: one is the regular summary of the main text in the chunk, while the other includes all surprising facts that differ from the main text.
The retriever architecture is illustrated in \Cref{fig:overview2}.

Building upon the idea that the informational value of a communicated message depends on the degree of surprise in its contents~\cite{el2011network}, the inclusion of surprising information,  distinct from the main text, will also provide valuable insights to LLMs when handling long contexts. 
After generating summary texts and surprising information from each chunk, we group similar texts using Gaussian Mixture Models, as employed in RAPTOR (refer to~\cite{sarthi2024raptor} for more details). However, we  only group the summaries without the surprising information.

All texts are embedded for searching and clustering using SBERT, a BERT-based encoder (multi-qa-mpnet-base-cos-v1)~\cite{reimers2019sentence}. These summarized texts are then re-embedded, and the cycle of embedding, clustering, and summarization continues until further clustering becomes impractical, resulting in a structured, multi-layered tree representation of the raw data. For querying within this tree, similar to RAPTOR, we use a collapsed tree strategy that disregards the tree structure and directly traverses all the nodes as shown in \Cref{fig:overview2}.

For inner-loop query part, as shown in Fig.~\ref{fig:overview}, we use an LLM as the answer model to generate the final answer. This model can be the same as the summary model or a separate one. We create an additional area called Short-Term Memory (STM), which stores texts up to the answer model maximum output length. The STM is initially empty. Each time, the answer model generates an answer based on the user's query, information retrieved by the retriever, and the previous texts in the STM. Then the answer will be stored in the STM. The retriever then uses the contents of the STM along with the user's query to retrieve new information from the raw data. Once new information is retrieved, the answer model generates new texts and stores them in the STM. This process is repeated in the inner-loop query until the texts in the STM converge and stop changing, or until the query limit is reached. Finally, the STM texts are returned to the user as the final answer.

\section{Experiments}

We evaluate ILM-TR’s long-context performance using two benchmarks: M-NIAH~\cite{kamradt2024needle} and BABILong~\cite{kuratov2024babilong}. For both the summary and answer model inference, we utilize Meta-Llama-3-70B with llama.cpp~\cite{llama_cpp}, quantized using Q4\_K\_M due to hardware limitations. We do not employ smaller LLMs, such as the 8B or 7B models, as their instruction-following capabilities were found to be inadequate in our tests: they consistently failed to follow the summarization prompts correctly. We also set the maximum number of inner-loop queries to 5. All tests were conducted on a machine running Ubuntu 22.04, equipped with an Intel Xeon Gold 6242 processor and four NVIDIA Tesla V100 32GB GPUs. Details of all prompts and parameters are provided in \Cref{sec:appendix.prompt,sec:appendix.parameter}.

In the M-NIAH test, several sentences are inserted into a specific area of a given long context. The question is related to all the inserted sentences, and the model is expected to retrieve all necessary information across these sentences. For example, we use three sentences: `Figs are one of the secret ingredients needed to build the perfect pizza', `Prosciutto is one of the secret ingredients needed to build the perfect pizza' and `Goat cheese is one of the secret ingredients needed to build the perfect pizza'. The question would then be `What is the first letter of each secret ingredient needed to build the perfect pizza?'. The BABILong test is similar to the M-NIAH test but involves sentences with more complex logical relationships. For instance, it may include sentences like `The apple is in the bathroom' and `Jack takes the apple to the kitchen' The question in this case would be: `Where is the apple before kitchen?'.

\begin{figure*}[t!]
\centering
\begin{subfigure}[b]{0.48\textwidth}
\centering
\includegraphics[width=\textwidth]{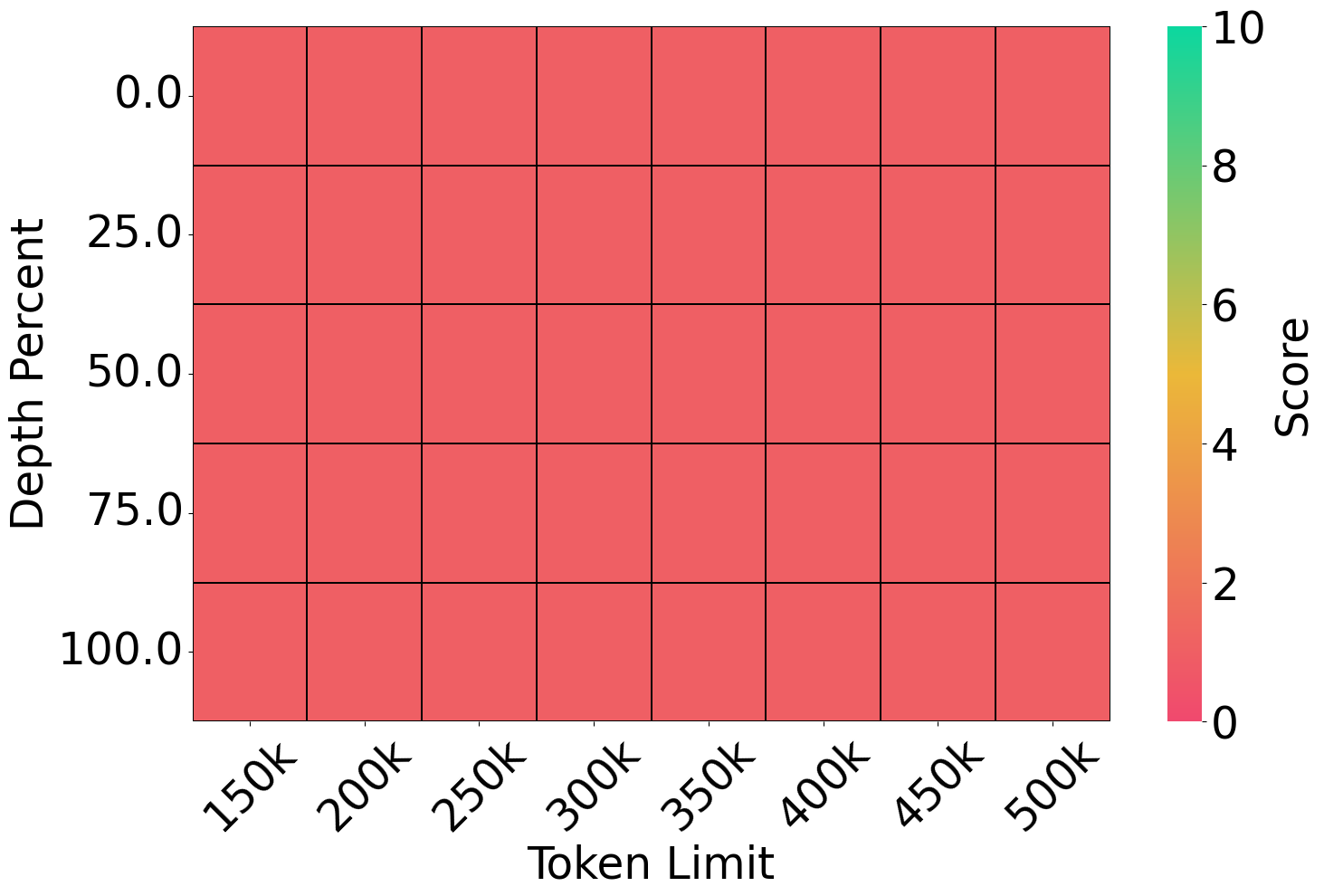} 
\caption{RAPTOR with Meta-Llama-3-70B}
\label{fig:NIAH.baseline}
\end{subfigure}
\hfill
\begin{subfigure}[b]{0.48\textwidth}
\centering
\includegraphics[width=\textwidth]{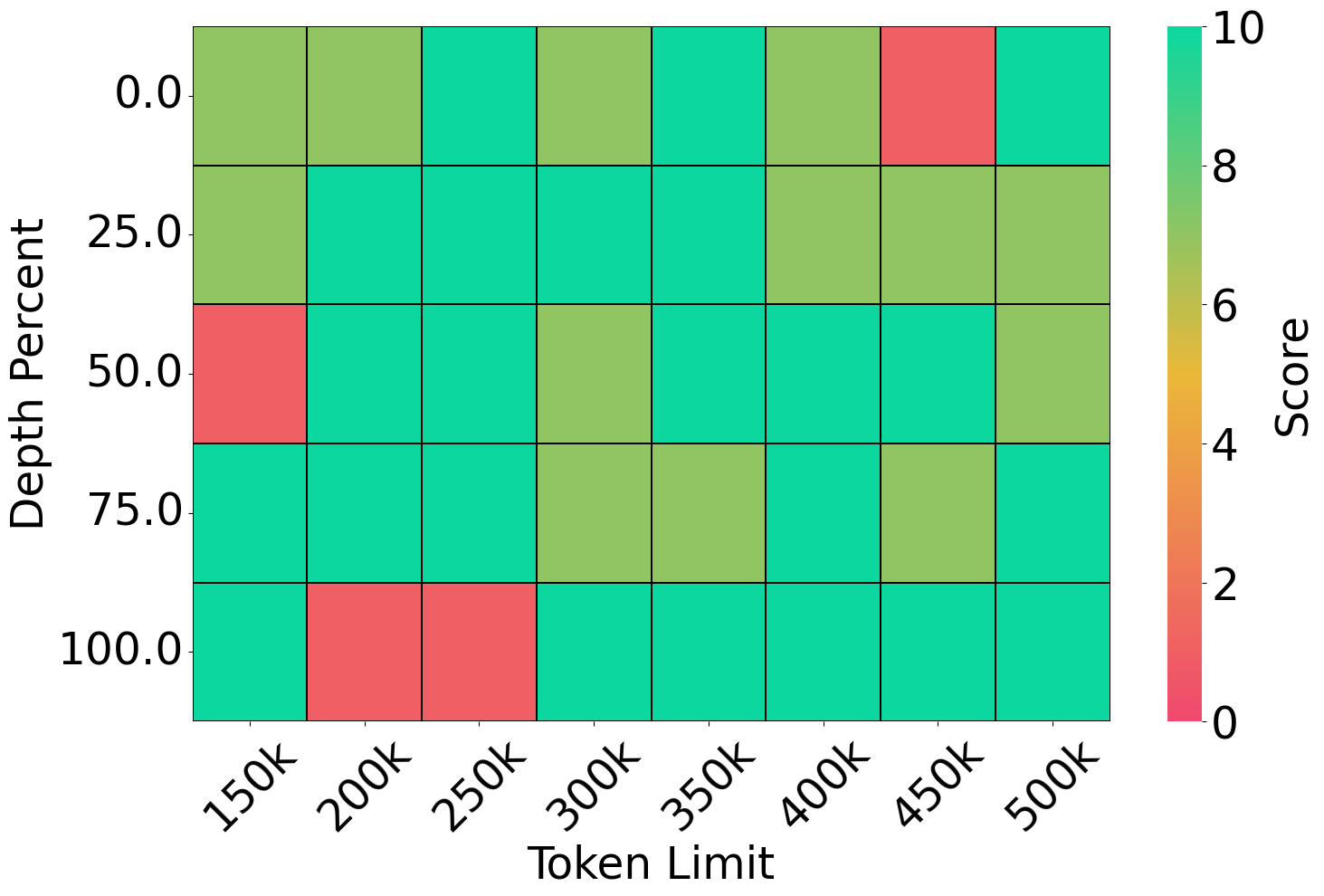} 
\caption{ILM-TR without inner-loop query
}
\label{fig:NIAH.noloop}
\end{subfigure}
\begin{subfigure}[b]{\textwidth}
\centering
\includegraphics[width=\textwidth]{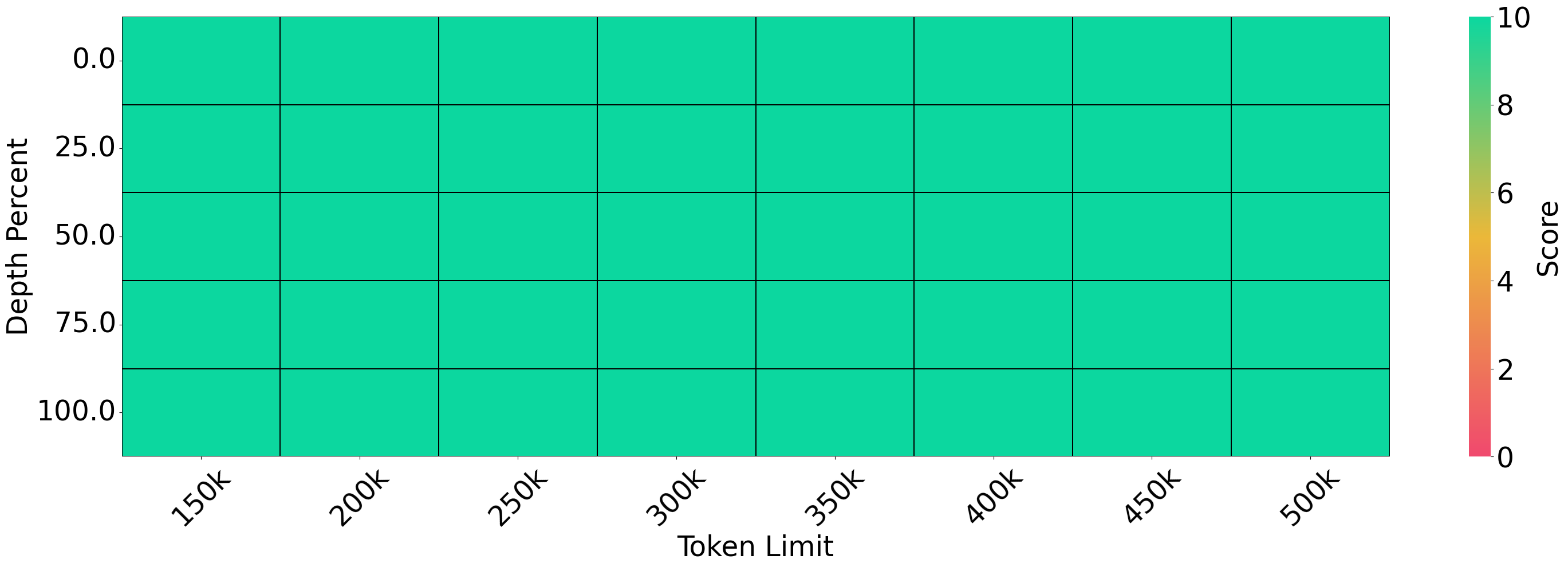} 
\caption{ILM-TR}
\label{fig:NIAH.ILMTR}
\end{subfigure}
\caption{M-NIAH test: no keywords found (score 1, red), one keyword found (score 3, orange), two keywords found (score 7, yellow), and all three keywords found (score 10, green). Token lengths range from 150k to 500k. Depth percent represents the average positions of the inserted sentences within the long text, where 0\% indicates the beginning of the text and 100\% indicates the end.
}
\label{fig:NIAH}
\end{figure*}

We present the M-NIAH and BABILong test results in \Cref{fig:NIAH} and \Cref{fig:BABILong}(\Cref{sec:appendix}). We tested RAPTOR as the baseline method, and our ILM-TR method with two settings, with token lengths ranging from 150k to 500k. There are three inserted sentences for M-NIAH test. Each testcase has four possible score levels: no keywords found (score 1, red), one keyword found (score 3, orange), two keywords found (score 7, yellow), and all keywords found (score 10, green). \Cref{fig:NIAH} demonstrates that incorporating surprising information can significantly improve the model’s performance in the M-NIAH test. However, there are still some cases where ILM-TR without the inner-loop query cannot retrieve all the keywords in a single query. The ILM-TR method, with inner-loop query capabilities, shows its ability to locate all the keywords in the M-NIAH test. In the BABILong test, our ILM-TR method also shows significant improvements compared to the baseline method. We show an simple example for BABILong with our ILM-TR model in \Cref{test:example}.

However, there are some shortcomings with ILM-TR. First, the inner-loop process requires several iterations, which increases the overall time consumption of query processing. Second, since we incorporate surprising information during summarization, the summary model must be capable of following complex instructions, which led us to choose a larger model with slower inference performance. 

\section{Conclusion and Future Work}

We introduce a novel approach, Inner Loop Memory Augmented Tree Retrieval (ILM-TR), which incorporates inner-loop queries based not only on the initial query but also on intermediate findings. During inference, ILM-TR retrieves information from the RAG system. Based on the retrieved information, ILM-TR generates text that is stored in Short-Term Memory (STM), which is then used to formulate subsequent queries. This retrieval process is repeated until the text in STM converges. Our experiments demonstrate that retrieval with STM offers improvements over traditional retrieval-augmented LLMs in the M-NIAH and BABILong test. And since the answers for M-NIAH tests are known, in future work, we can explore fine-tuning the answer model output based on the intermediate results from STM and the reference answer. This could potentially improve model's active search capabilities with RAG system.

\bibliographystyle{aaai25}
\bibliography{strings,myref}


\newpage
\appendix

\section{Appendix / supplemental material}
\label{sec:appendix}

\begin{figure*}[t!]
\centering
\begin{subfigure}[b]{0.48\textwidth}
\centering
\includegraphics[width=\textwidth]{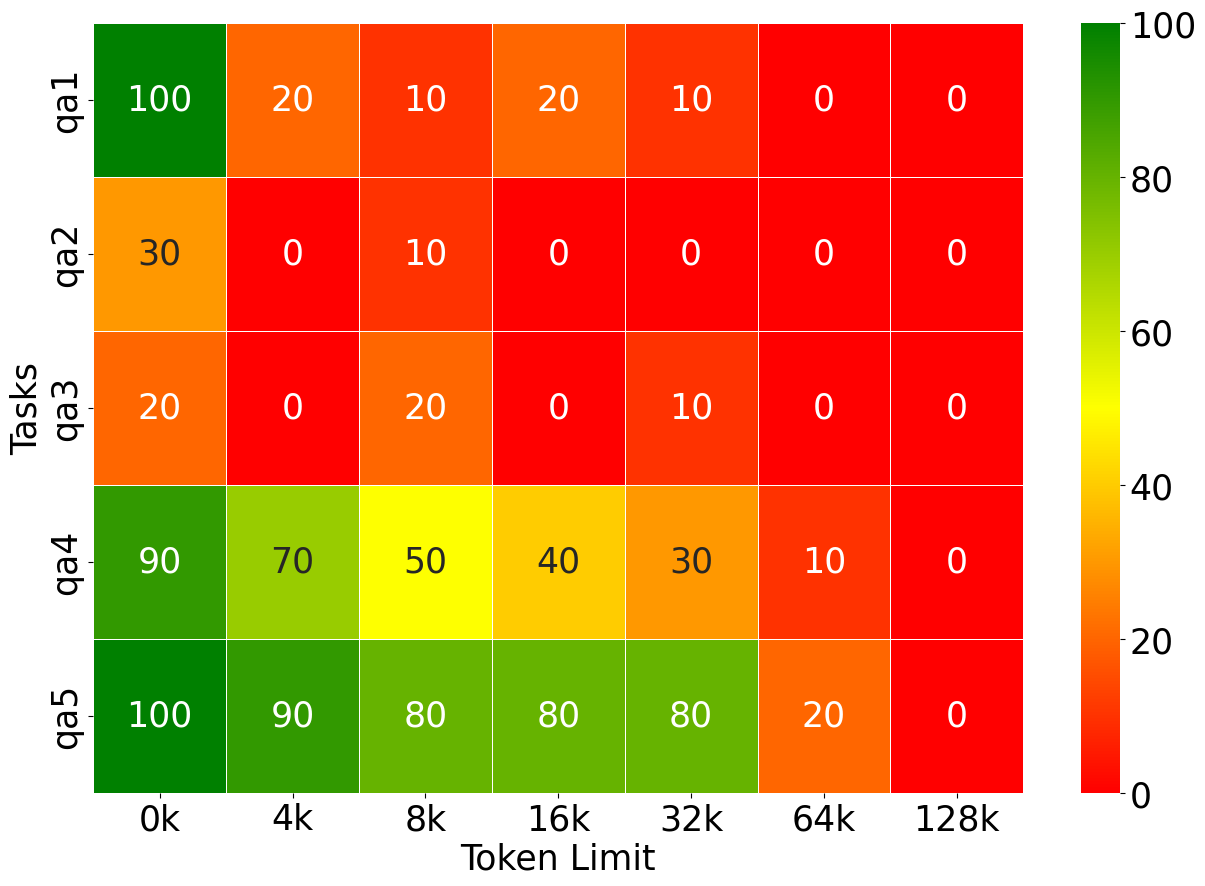} 
\caption{RAPTOR with Meta-Llama-3-70B}
\label{fig:BABILong.baseline}
\end{subfigure}
\hfill
\begin{subfigure}[b]{0.48\textwidth}
\centering
\includegraphics[width=\textwidth]{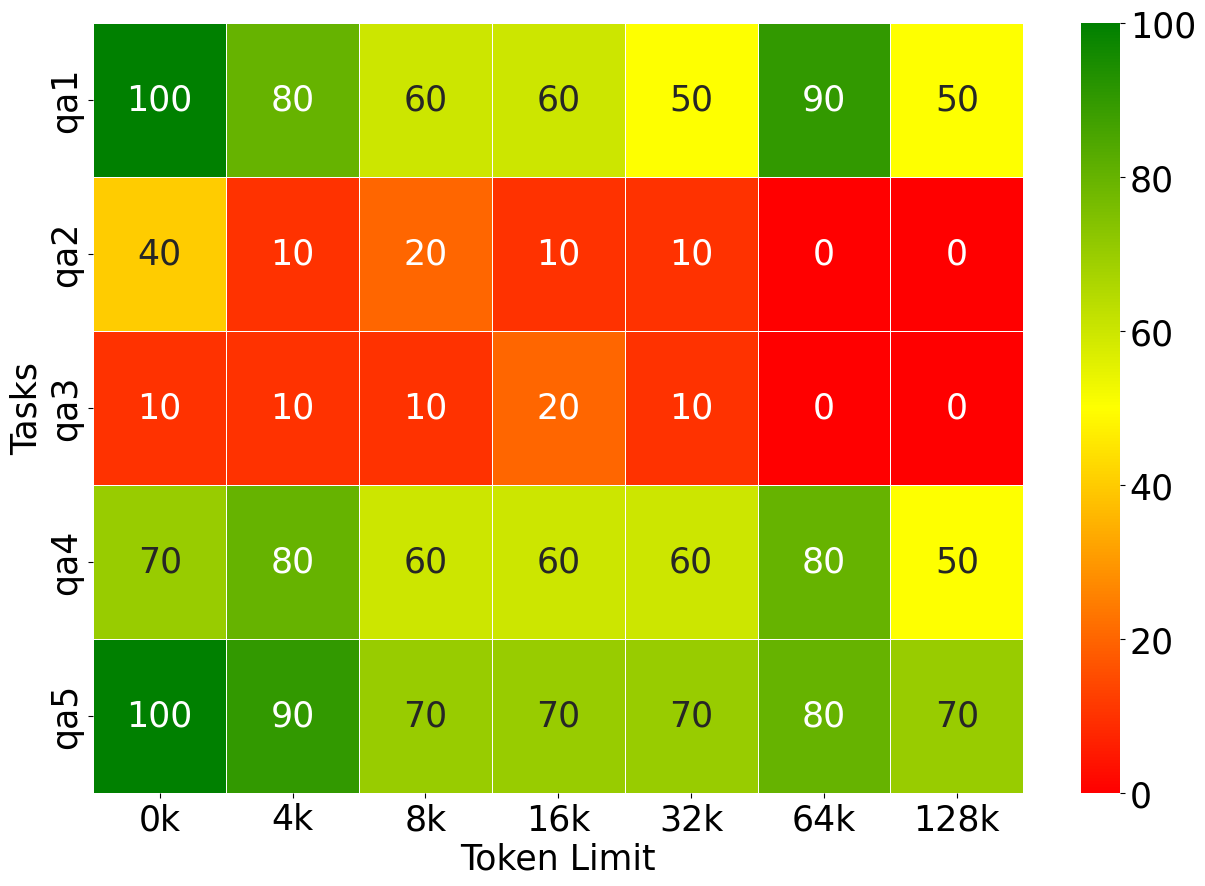} 
\caption{ILM-TR without inner-loop query
}
\label{fig:BABILong.noloop}
\end{subfigure}
\begin{subfigure}[b]{0.5\textwidth}
\centering
\includegraphics[width=\textwidth]{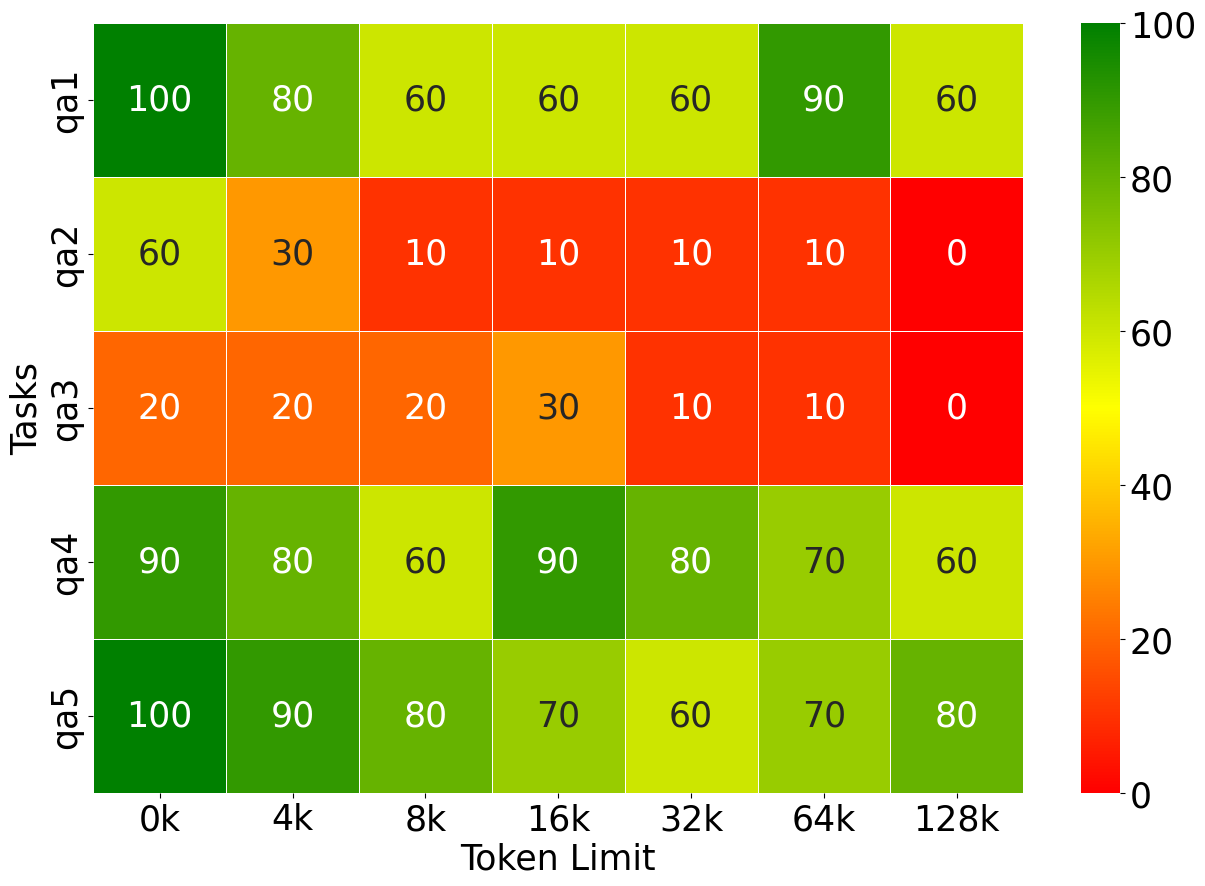} 
\caption{ILM-TR}
\label{fig:BABILong.ILMTR}
\end{subfigure}
\caption{BABILong test: Due to hardware limitations and the size of the LLM model, we only tested 10 testcases for each test setting. In this study, we evaluated tasks qa1 to qa5 with token lengths ranging from 0k to 128k. We encourage readers to refer to~\cite{kuratov2024babilong} for further details.
}
\label{fig:BABILong}
\end{figure*}

\subsection{Prompt}
\label{sec:appendix.prompt}

\subsubsection{Baseline Prompt}
\begin{quote}
\textbf{Summary Model} 

\textbf{[SYSTEM]}: You are a reader who can summarize the given text while including important details. Do not provide any comments, just give the summary.

\textbf{[USER]}: Write a summary of the following context, just including the most important details: \{context\}
\end{quote}

\begin{quote}
\textbf{Answer Model} 

\textbf{[SYSTEM]}: You are Question Answering Portal.

\textbf{[USER]}: Given Context: \{Retrieved Info\} Give the best full answer to question \{User's Query\}
\end{quote}

\subsubsection{ILM-TR Prompt}

\begin{quote}
\textbf{Summary Model} 

\textbf{[SYSTEM]}: I will give you context. Most of it could be about the same things, but there may be some abnormal information. An surprising sentence is not related to most of the other content. You should summarize the context with the necessary information and also include any surprising information you think. Don't return any unrelated words. 

Always and only return your answer with the following format, just list the fact based on given text, no comments, use different sentences to describe different facts:

(Summary): Your Summary

(Surprise): Surprising Information

\textbf{[USER]}: \{context\}
\end{quote}

\begin{quote}
\textbf{Answer Model without inner-loop} 

\textbf{[SYSTEM]}: You are Question Answering Portal.

\textbf{[USER]}: Given Context: \{Retrieved Info\} Give the best full answer to question \{User's Query\}
\end{quote}

\begin{quote}
\textbf{Answer Model with inner-loop} 

\textbf{[SYSTEM]}: You will be given some Retrieved Info and memory, and you will use this information to answer a question. If you can't answer the question, you can write something related to the question to help others answer it. \\
(Retrieved Info): \\
a2 = 1\\
(Memory):\\
a1 = a2+a3 \\
(Question): \\
What is the value of a1 \\
(Your Output): \\
a1 = a2 + a3. a2=1. We need to find the value of a3.

Keep your output short and don't return any unrelated words. 

\textbf{[USER]}: \{context\} (Question): \{User’s Query\}
\end{quote}

\subsection{BABILong Example with ILM-TR}
\label{test:example}

\subsubsection{Example (qa3, 128k)}
\begin{quote}
\textbf{[Needles]}:\\
Daniel grabbed the milk. \\
Mary picked up the apple. \\
Sandra went back to the hallway. \\
Daniel journeyed to the hallway. \\
John moved to the bedroom. \\
John went to the bathroom. \\
Daniel discarded the milk there. \\
Mary moved to the kitchen. \\
Mary journeyed to the office. \\
Daniel got the milk. \\
John moved to the garden. \\
Sandra travelled to the kitchen. \\
Mary put down the apple. \\
John took the football. \\
\textbf{[Question]}: Where was the apple before the office? \\
\textbf{[True Answer]}: kitchen \\
\end{quote}

\subsubsection{ILM-TR Inner-loop Outputs}
We calculate the Longest Common Subsequence (LCS) to determine whether the answer has converged.
\begin{quote}
\textbf{[Round 1]}:\\
The apple was at office. We need to find where the apple was before the office. \\
\textbf{[Round 2]}:\\
The apple was at office. Mary put down the apple at office. We need to determine where Mary was before she placed the apple down. \\
\textbf{[Round 3]}:\\
The apple was at office. Mary put down the apple at office, but before that, she was in the kitchen.\\
\textbf{[Round 4]}:\\
Mary put down the apple at office, but before that, she was in the kitchen. The best answer to the question ""Where was the apple before the office?"" is:

The kitchen.\\
\textbf{[Round 5]}:\\
Based on the given context, the best answer to the question ""Where was the apple before the office?"" is:

The kitchen.\\
\end{quote}

\subsection{Model and Retriever Parameters}
\label{sec:appendix.parameter}

For parameters not mentioned here, the values are defualt values.

\begin{longtable}{ll}
\toprule
\textbf{Parameter Category} & \textbf{Parameter Value} \\
\midrule
\multicolumn{2}{l}{\textbf{Answer Model}} \\
\quad temperature & 0 \\
\quad frequency\_penalty & 1.2 \\
\quad max\_tokens & 200 \\
\midrule
\multicolumn{2}{l}{\textbf{Summary Model}} \\
\quad repeat\_last\_n & 256 \\
\quad repeat\_penalty & 1.18 \\
\quad penalize\_nl & False \\
\quad presence\_penalty & 0 \\
\quad min\_p & 0.05 \\
\quad n\_predict & 1055 \\
\quad n\_probs & 0 \\
\quad mirostat & 0 \\
\quad mirostat\_eta & 0.1 \\
\quad mirostat\_tau & 5 \\
\quad tfs\_z & 1 \\
\quad top\_k & 40 \\
\quad top\_p & 0.95 \\
\quad typical\_p & 1 \\
\quad frequency\_penalty & 0 \\
\quad temperature & 0.2 \\
\midrule
\multicolumn{2}{l}{\textbf{Retriever}} \\
\quad tb\_max\_tokens (each chunk max token) & 600 \\
\quad tb\_summarization\_length (output max token) & 300 \\
\bottomrule
\end{longtable}


\end{document}